\PassOptionsToPackage{unicode}{hyperref}
\PassOptionsToPackage{hyphens}{url}
\PassOptionsToPackage{dvipsnames,svgnames,x11names}{xcolor}
\documentclass[
]{article}

\usepackage{amsmath,amssymb}
\usepackage{iftex}
\ifPDFTeX
  \usepackage[T1]{fontenc}
  \usepackage[utf8]{inputenc}
  \usepackage{textcomp} 
\else 
  \usepackage{unicode-math}
  \defaultfontfeatures{Scale=MatchLowercase}
  \defaultfontfeatures[\rmfamily]{Ligatures=TeX,Scale=1}
\fi
\usepackage{lmodern}
\ifPDFTeX\else  
\fi
\IfFileExists{upquote.sty}{\usepackage{upquote}}{}
\IfFileExists{microtype.sty}{
  \usepackage[]{microtype}
  \UseMicrotypeSet[protrusion]{basicmath} 
}{}
\makeatletter
\@ifundefined{KOMAClassName}{
  \IfFileExists{parskip.sty}{%
    \usepackage{parskip}
  }{
    \setlength{\parindent}{0pt}
    \setlength{\parskip}{6pt plus 2pt minus 1pt}}
}{
  \KOMAoptions{parskip=half}}
\makeatother
\usepackage{xcolor}
\ifx\paragraph\undefined\else
  \let\oldparagraph\paragraph
  \renewcommand{\paragraph}[1]{\oldparagraph{#1}\mbox{}}
\fi
\ifx\subparagraph\undefined\else
  \let\oldsubparagraph\subparagraph
  \renewcommand{\subparagraph}[1]{\oldsubparagraph{#1}\mbox{}}
\fi

\providecommand{\tightlist}{%
  \setlength{\itemsep}{0pt}\setlength{\parskip}{0pt}}\usepackage{longtable,booktabs,array}
\usepackage{calc} 
\usepackage{etoolbox}
\makeatletter
\patchcmd\longtable{\par}{\if@noskipsec\mbox{}\fi\par}{}{}
\makeatother
\IfFileExists{footnotehyper.sty}{\usepackage{footnotehyper}}{\usepackage{footnote}}
\makesavenoteenv{longtable}
\usepackage{graphicx}
\makeatletter
\def\maxwidth{\ifdim\Gin@nat@width>\linewidth\linewidth\else\Gin@nat@width\fi}
\def\maxheight{\ifdim\Gin@nat@height>\textheight\textheight\else\Gin@nat@height\fi}
\makeatother
\setkeys{Gin}{width=\maxwidth,height=\maxheight,keepaspectratio}
\makeatletter
\def\fps@figure{htbp}
\makeatother
\newlength{\cslhangindent}
\newlength{\csllabelwidth}
\newlength{\cslentryspacingunit} 
\newenvironment{CSLReferences}[2] 
 {
  \setlength{\parindent}{0pt}
  \ifodd #1
  \let\oldpar\par
  \def\par{\hangindent=\cslhangindent\oldpar}
  \fi
  \setlength{\parskip}{#2\cslentryspacingunit}
 }%
 {}
\usepackage{calc}

\usepackage{arxiv}
\usepackage{orcidlink}
\usepackage{amsmath}
\usepackage[T1]{fontenc}
\makeatletter
\@ifpackageloaded{caption}{}{\usepackage{caption}}
\AtBeginDocument{%
\ifdefined\contentsname
  \renewcommand*\contentsname{Table of contents}
\else
  \newcommand\contentsname{Table of contents}
\fi
\ifdefined\listfigurename
  \renewcommand*\listfigurename{List of Figures}
\else
  \newcommand\listfigurename{List of Figures}
\fi
\ifdefined\listtablename
  \renewcommand*\listtablename{List of Tables}
\else
  \newcommand\listtablename{List of Tables}
\fi
\ifdefined\figurename
  \renewcommand*\figurename{Figure}
\else
  \newcommand\figurename{Figure}
\fi
\ifdefined\tablename
  \renewcommand*\tablename{Table}
\else
  \newcommand\tablename{Table}
\fi
}
\@ifpackageloaded{float}{}{\usepackage{float}}
\floatstyle{ruled}
\@ifundefined{c@chapter}{\newfloat{codelisting}{h}{lop}}{\newfloat{codelisting}{h}{lop}[chapter]}
\floatname{codelisting}{Listing}

\makeatother
\makeatletter
\@ifpackageloaded{caption}{}{\usepackage{caption}}
\@ifpackageloaded{subcaption}{}{\usepackage{subcaption}}
\makeatother
\makeatletter
\@ifpackageloaded{tcolorbox}{}{\usepackage[skins,breakable]{tcolorbox}}
\makeatother
\makeatletter
\@ifundefined{shadecolor}{\definecolor{shadecolor}{rgb}{.97, .97, .97}}
\makeatother
\ifLuaTeX
  \usepackage{selnolig}  
\fi
\IfFileExists{bookmark.sty}{\usepackage{bookmark}}{\usepackage{hyperref}}
\IfFileExists{xurl.sty}{\usepackage{xurl}}{} 
\hypersetup{
  pdftitle={AI-assisted coding: Experiments with GPT-4},
  pdfauthor={Russell A. Poldrack; Thomas Lu; Gaper Begu},
  pdfkeywords={artificial intelligence, software
engineering, reproducibility},
  colorlinks=true,
  linkcolor={blue},
  filecolor={Maroon},
  citecolor={Blue},
  urlcolor={Blue},
  pdfcreator={LaTeX via pandoc}}

\newcommand{\runninghead}{ }
\renewcommand{\runninghead}{Coding with GPT-4 }
\title{AI-assisted coding: Experiments with GPT-4}
\author{
\textbf{Russell A. Poldrack}~\orcidlink{0000-0001-6755-0259}\\Department
of Psychology\\Stanford University\\Stanford,
CA,\ 94305\\\href{mailto:russpold@stanford.edu}{russpold@stanford.edu}\\\\\\
\textbf{Thomas Lu}\\Department of Linguistics\\University of
California\\Berkeley, CA,\ 94720-2650\\\\\\\\
\textbf{Ga\v{s}per
Begu\v{s}}~\orcidlink{0000-0002-6459-0551}\\Department of
Linguistics\\University of California\\Berkeley,
CA,\ 94720-2650\\\href{mailto:begus@berkeley.edu}{begus@berkeley.edu}}
\date{}
\begin{document}
\maketitle
\begin{abstract}
Artificial intelligence (AI) tools based on large language models have
acheived human-level performance on some computer programming tasks. We
report several experiments using GPT-4 to generate computer code. These
experiments demonstrate that AI code generation using the current
generation of tools, while powerful, requires substantial human
validation to ensure accurate performance. We also demonstrate that
GPT-4 refactoring of existing code can significantly improve that code
along several established metrics for code quality, and we show that
GPT-4 can generate tests with substantial coverage, but that many of the
tests fail when applied to the associated code. These findings suggest
that while AI coding tools are very powerful, they still require humans
in the loop to ensure validity and accuracy of the results.
\end{abstract}
{\bfseries \emph Keywords}
\def\sep{\textbullet\ }
artificial intelligence \sep software engineering \sep 
reproducibility

\ifdefined\Shaded\renewenvironment{Shaded}{\begin{tcolorbox}[sharp corners, interior hidden, breakable, borderline west={3pt}{0pt}{shadecolor}, boxrule=0pt, enhanced, frame hidden]}{\end{tcolorbox}}\fi

\hypertarget{sec-intro}{%
\section{Introduction}\label{sec-intro}}

Recent developments in artificial intelligence, particularly through
large language models, have enabled the automated generation of computer
code (Chen et al. 2021; Bubeck et al. 2023). In particular, GPT-4 has
enabled human-level performance on a set of coding challenges that are
outside of the training set of the model (Bubeck et al. 2023). In
addition, automated coding assistants (particularly Github Copilot) have
become integrated into commmon devlopment environments and are widely
used, with some evidence that they can signficantly improve coding
productivity. The performance of these models is also raising important
questions regarding coding education, given that the current models can
easily complete most coding problem sets using in introductory
programming courses (Finnie-Ansley et al. 2022).

In the present paper we explore some of the implications of AI-assisted
coding using GPT-4, in a more qualitative way than previous benchmarking
assessments. First we examine the experience of interactive coding and
debugging using the ChatGPT interface to GPT-4 on a set of data science
coding problems. This experiment is meant to approximate the experience
of a researcher with minimal expertise in prompt engineering, assessing
the success and amount of effort required to perform these coding tasks.
Second, we assess the ability of GPT-4 (using the OpenAI API) to
refactor and improve the quality of existing code. This experiment is
meant to assess the degree to which AI coding assistants might improve
coding quality when used by researcers. Third, we assess the ability of
GPT-4 to write tests for its own code, using a set of test prompts from
several scientific domains. We conclude with an overall assessment of
the utility of AI coding assistants for scientific researchers.

A fully reproducible workflow for this manuscript is available at
\url{https://github.com/poldrack/ai-coding-experiments}.

\hypertarget{coding-with-gpt-4}{%
\section{Coding with GPT-4}\label{coding-with-gpt-4}}

Our first set of experiments examined the ability of GPT-4 (via the
ChatGPT interface) to generate usable code for a set of data science
problems. The prompts were generated manually (by author RP) and are
listed in Appendix 1. Each prompt was submitted in a separate chat
session; the human examined the resulting code, and issued additional
prompts to try to fix problems. If the human was not able to obtain
working code within about 5 minutes of effort or less, the problem was
deemed unsolved. The results of this experiment are primarily
qualitiative and subjective, but are meant to index the degree to which
GPT-4 is a useful tool for a researcher with minimal prompt engineering
skills.

\begin{figure}

{\centering \includegraphics{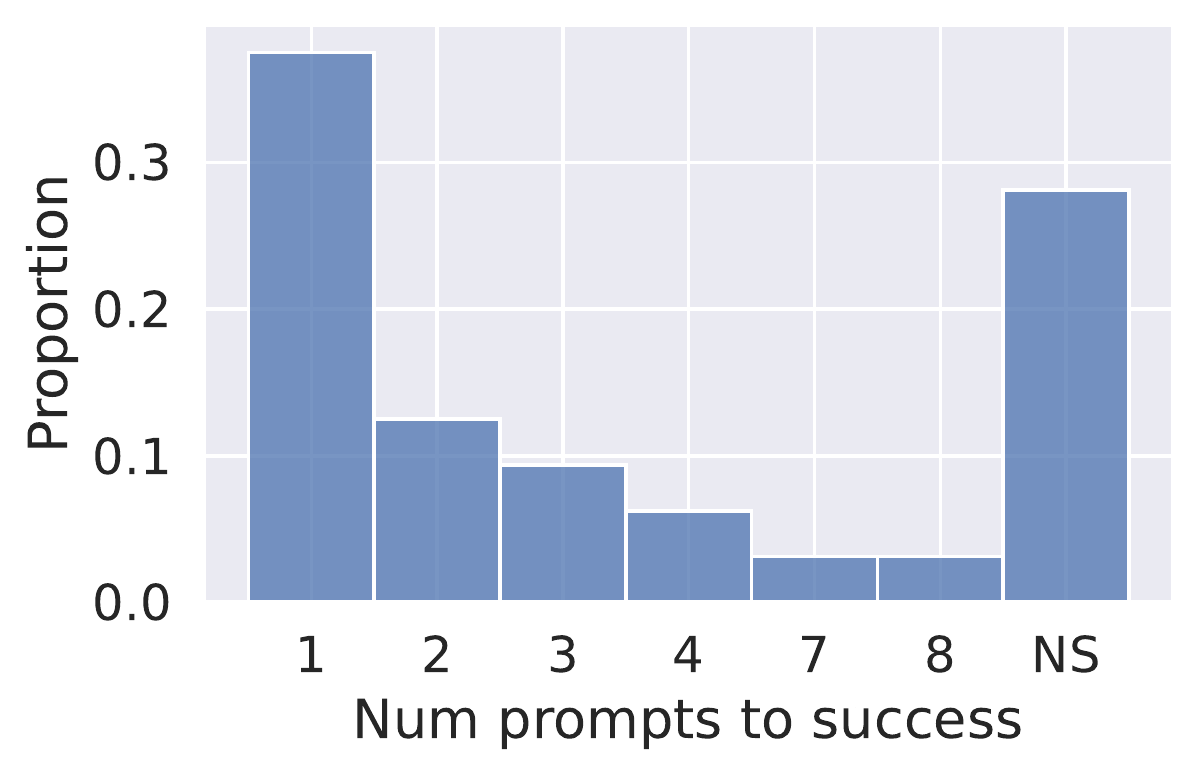}

}

\caption{\label{fig-prompts-plot}Proportion of successful code outcomes
as a function of number of prompts. NS: not successful.}

\end{figure}

Figure~\ref{fig-prompts-plot} shows the proportion of successful
outcomes as a function of the number of prompts required. 72\% (23/32))
of attempts were successful in relatively quickly solving the problem;
37.5\% (12/32) were successful on the first prompt. In cases where
additional prompting was required, a common problem was the use of
outdated functions or datasets from Python libraries.

The causes of unsuccessful outcomes were varied. In some cases it was
due to the outdated nature of the GPT-4 training data. For example, in
one case (prompt \#12) ChatGPT could not successfully implement a
solution that was compatible with the latest version of PyTorch. In
another case (prompt \#27), the labels used to query the NOAA API were
incorrect, and the correct labels were not easily identifable upon
further examination. In other cases, it was not immediately clear what
was wrong with the code, and the researcher did not dig deeply enough to
identify the root cause.

One of the examples highlights the contining challenges that ChatGPT has
with mathematical processing (as outlined by Bubeck et al. (2023)).
Prompt \#4 asked ChatGPT to generate code to simulate a drift diffusion
model (a common model for response times in cognitive psychology) and
then fit the model parameters using the EZ-diffusion model (Wagenmakers,
Van Der Maas, and Grasman 2007), which is a closed-form solution to
estimating these parameters using response time and accuracy statistics.
The initial code generated by ChatGPT attempted to fit a diffusion model
through numerical optimization. After being prompted to generate a
closed-form solution based on the original paper, ChatGPT did so, but
the formula bore little resemblance to the actual formula from the
paper. This is an example of a ``hallucination'' which is commonly seen
with LLMs (Ji et al. 2023), and highlights the ongoing need for
automatically generated code to be validated by a human.

Another example also highlights the need for sophisticated domain
knowledge in assessing the outputs of ChatGPT. Prompt \#18 asked ChatGPT
to implement a \emph{hurdle model}, which is a statistical model for
zero-inflated count data that combines a binary model with count model
using a truncated distribution. In general, this model is fit by
performing maximum likelihood estimation on the combined likelihoods of
the binary and count models. ChatGPT generated a solution that
separately estimated a binary model and a count model, and then combined
the predictions from the two models; this incorrect approach can be
found in a number of examples from Github. This model fit the test data
nearly as well as a reference implementation of the hurdle model
(impmented in R), but is incorrect in comparison to the reference
implementation. This highlights the need for close attention to detail
in the implementation of any numerical methods, as incorrect
implementations present on Github can result in incorrect outcomes.

\hypertarget{refactoring-code-using-gpt4}{%
\subsection{Refactoring code using
GPT4}\label{refactoring-code-using-gpt4}}

The quality of research code is important for scientific reproducibility
and transparency, as well as for code reusability and maintainability.
In our initial explorations of code generated by GPT-4, we noted that
the automatically generated code appeared to be substantially more
readable than research code that we have encountered (or written) in the
past. This led us to ask whether GPT-4 could improve the quality of
existing code through \emph{refactoring} (Fowler 2019), by which we mean
modifying code to make it more readable or maintainable without changing
its behavior.

To assess this, we downloaded more than 2000 examples of Python code
from Github using the Githb Code Search API. Only one code file was
allowed from any single repository. We further filtered these files,
based in part on the criteria used by Chen et al. (2021) to select code
for training of the OpenAI Codex model. Exclusion criteria included:

\begin{itemize}
\tightlist
\item
  Presence of non-Python code (as guessed by \texttt{guesslang.Guess()})
\item
  Presence of non-English language in the code (according to
  \texttt{pycld2.detect()})
\item
  Presence of potential markers of automatic generation (e.g.~strings
  such as ``autogenerated'', ``generated by Django'', etc)
\item
  Presence of potential obfuscation (the ``if x - y:'' pattern)
\item
  Lack of at least one function definition
\item
  Number of GPT tokens greater than 2500 or less than 200
\item
  Maximum line length \textgreater{} 1000
\item
  Mean line length \textgreater{} 100
\item
  Maximum file size \textgreater{} 1MB
\end{itemize}

The 274 code files passing these criteria were submitted to further
analysis. Analysis of syntax errors and coding style was performed using
the \texttt{flake8} static code analysis package. Files were further
excluded on the basis of any syntax errors identified by flake8.

\begin{figure}

{\centering \includegraphics[width=0.6\textwidth,height=\textheight]{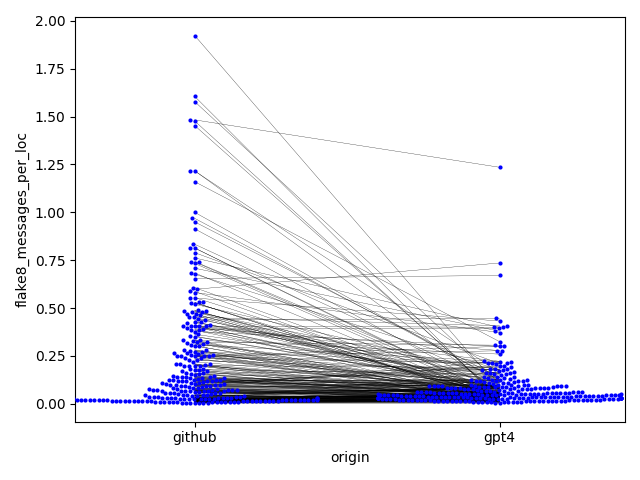}

}

\caption{\label{fig-messages-plot}Number of Flake8 messages (per line of
code) for original github files and refactored files.}

\end{figure}

\begin{figure}

{\centering \includegraphics{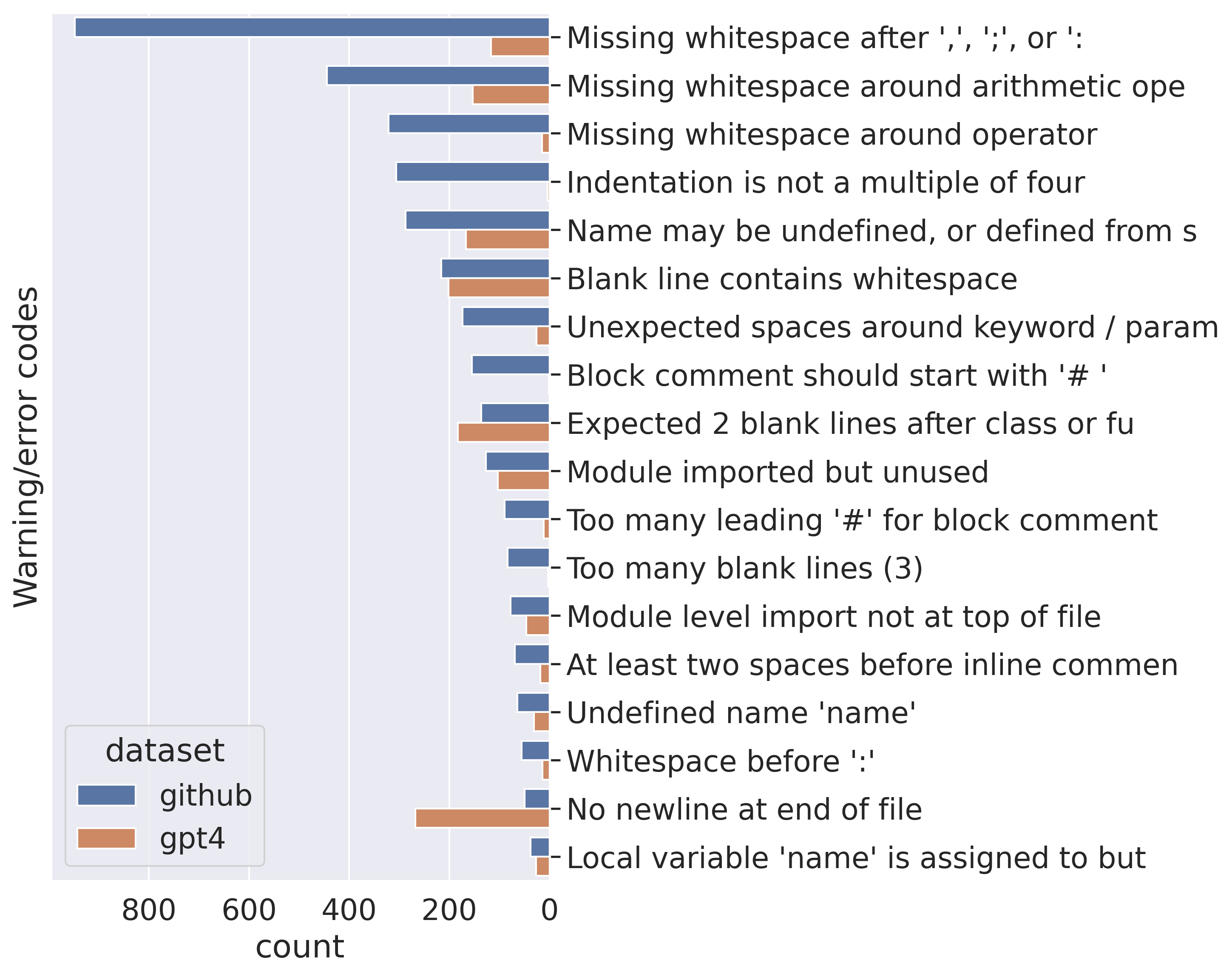}

}

\caption{\label{fig-errors-plot}Prevalence of individual Flake8
warning/error codes for original github files and refactored files.
Values are sorted by prevalence in the original github files.}

\end{figure}

The number of warning and error messages emitted by the flake8 linter
was substantially reduced for the refactored code compared to the
original (median 0.23 messages/line for original vs 0.09 messages/line
for refactored, Cohen's d = 0.50; see Figure~\ref{fig-messages-plot}).
While tools exist to perform automatic reformatting to ensure
standard-compliance, this shows that GPT-4 generates code that is
substantially more standards-compliant than the average programmer;
given that the files sampled from Github were heavily filtered, this
result is probably an underestimate compared to the broader population
of Python code on Github. Figure~\ref{fig-errors-plot} provides an
overview of which errors were most common in the original code and how
their prevalence changed after refactoring.

We further examined a set of code quality metrics, which were computed
using the \texttt{radon} Python package. Metrics extracted from these
outputs for further analysis included:

\begin{itemize}
\tightlist
\item
  Logical lines of code
\item
  Number of comments
\item
  Mean cyclomatic complexity (a measure of the number of execution
  paths)
\item
  Maintainability index (a holistic metric for code maintainability,
  based on a composite of several metrics including Halstead volume
  Halstead (1977), cyclomatic complexity, lines of code, and \% of
  comments)
\item
  Halstead ``difficulty'' (a metric of how difficult the code is to
  read, based on the number of distinct operators and the ratio of total
  number of operands to number of distinct operands)
\item
  Halstead ``bugs'' (a metric meant to estimate the number of bugs in
  delivered code)
\end{itemize}

\begin{figure}

\begin{minipage}[t]{0.50\linewidth}

{\centering 

\raisebox{-\height}{

\includegraphics{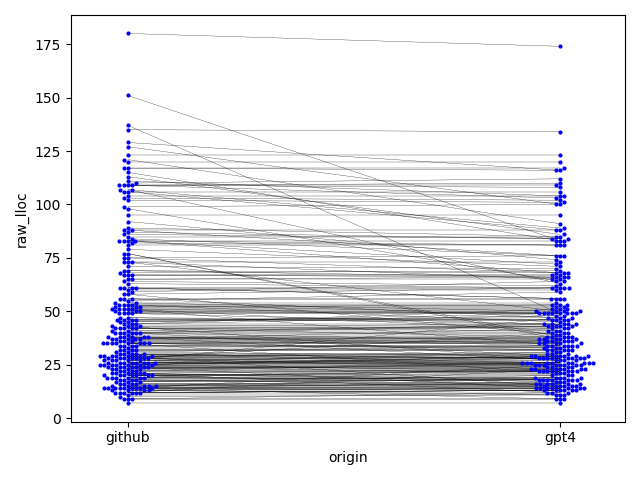}

}

}

\subcaption{\label{fig-lloc}Logical lines of code}
\end{minipage}%
\begin{minipage}[t]{0.50\linewidth}

{\centering 

\raisebox{-\height}{

\includegraphics{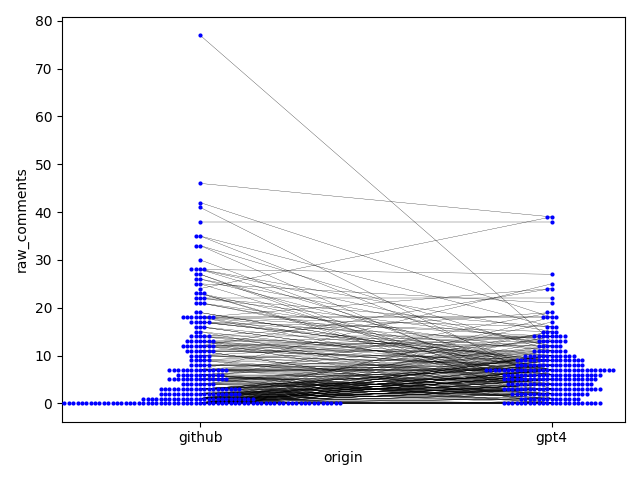}

}

}

\subcaption{\label{fig-comments}Comments}
\end{minipage}%
\newline
\begin{minipage}[t]{0.50\linewidth}

{\centering 

\raisebox{-\height}{

\includegraphics{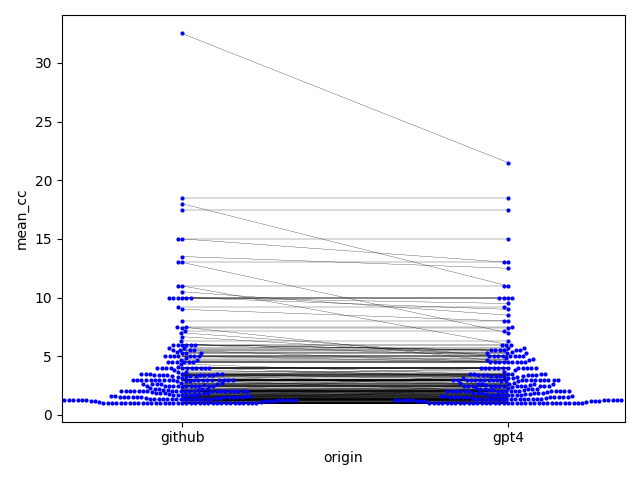}

}

}

\subcaption{\label{fig-cc}Mean cyclomatic complexity}
\end{minipage}%
\begin{minipage}[t]{0.50\linewidth}

{\centering 

\raisebox{-\height}{

\includegraphics{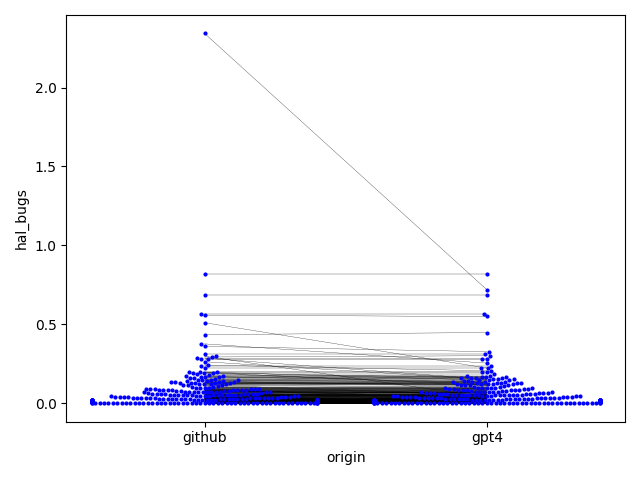}

}

}

\subcaption{\label{fig-bugs}Halstead bugs}
\end{minipage}%
\newline
\begin{minipage}[t]{0.50\linewidth}

{\centering 

\raisebox{-\height}{

\includegraphics{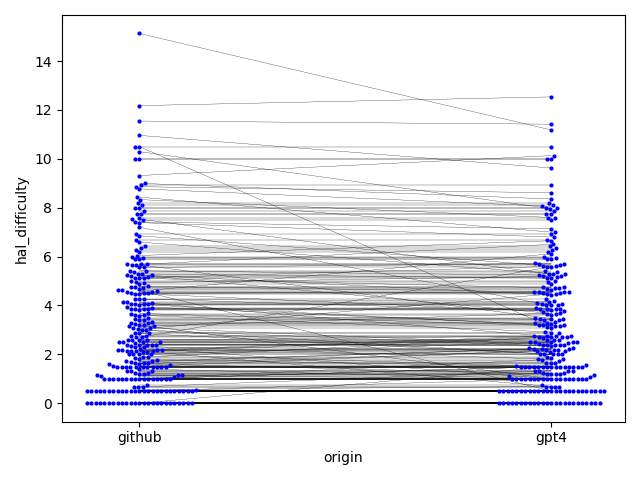}

}

}

\subcaption{\label{fig-diff}Halstead difficulty}
\end{minipage}%
\begin{minipage}[t]{0.50\linewidth}

{\centering 

\raisebox{-\height}{

\includegraphics{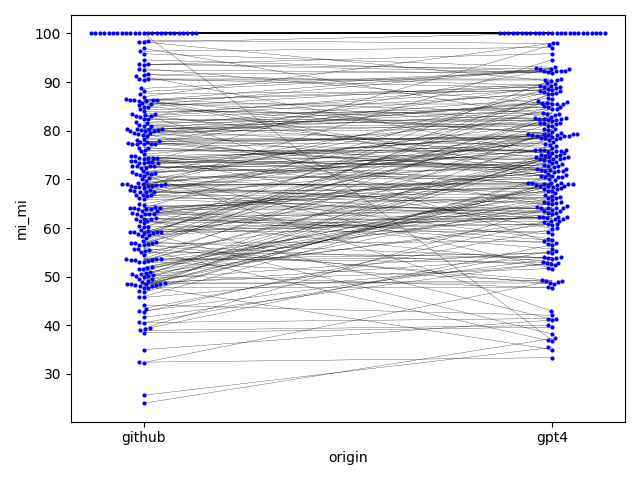}

}

}

\subcaption{\label{fig-mi}Maintainability index}
\end{minipage}%

\caption{\label{fig-metrics}Code quality metrics computed for each
original file and its refactored version.}

\end{figure}

Comparisons between metrics for the original and refactored code are
shown in Figure~\ref{fig-metrics}, and means, effect sizes, and p-values
for the comparison (using a paired t-test) are shown in
Table~\ref{tbl-metrics}. Each of these metrics differed between the
origin and refactored code (p \textless{} .05 after false discovery rate
correction across all hypotheses). However, the effect sizes were all in
the small to medium range, with Cohen's d values ranging from 0.13 to
0.33.

\hypertarget{tbl-metrics}{}
\begin{longtable}[]{@{}
  >{\raggedright\arraybackslash}p{(\columnwidth - 8\tabcolsep) * \real{0.2747}}
  >{\raggedleft\arraybackslash}p{(\columnwidth - 8\tabcolsep) * \real{0.2088}}
  >{\raggedleft\arraybackslash}p{(\columnwidth - 8\tabcolsep) * \real{0.1978}}
  >{\raggedleft\arraybackslash}p{(\columnwidth - 8\tabcolsep) * \real{0.1319}}
  >{\raggedleft\arraybackslash}p{(\columnwidth - 8\tabcolsep) * \real{0.1868}}@{}}
\caption{\label{tbl-metrics}Metric comparisons between original and
refactored code.}\tabularnewline
\toprule\noalign{}
\begin{minipage}[b]{\linewidth}\raggedright
\end{minipage} & \begin{minipage}[b]{\linewidth}\raggedleft
Median (github)
\end{minipage} & \begin{minipage}[b]{\linewidth}\raggedleft
Median (GPT-4)
\end{minipage} & \begin{minipage}[b]{\linewidth}\raggedleft
Cohens d
\end{minipage} & \begin{minipage}[b]{\linewidth}\raggedleft
P-value (FDR)
\end{minipage} \\
\midrule\noalign{}
\endfirsthead
\toprule\noalign{}
\begin{minipage}[b]{\linewidth}\raggedright
\end{minipage} & \begin{minipage}[b]{\linewidth}\raggedleft
Median (github)
\end{minipage} & \begin{minipage}[b]{\linewidth}\raggedleft
Median (GPT-4)
\end{minipage} & \begin{minipage}[b]{\linewidth}\raggedleft
Cohens d
\end{minipage} & \begin{minipage}[b]{\linewidth}\raggedleft
P-value (FDR)
\end{minipage} \\
\midrule\noalign{}
\endhead
\bottomrule\noalign{}
\endlastfoot
Maintainability index & 70.285 & 74.092 & 0.33 & \textless.001 \\
Halstead bugs & 0.081 & 0.068 & 0.13 & 0.045 \\
Halstead difficulty & 3.214 & 3.089 & 0.16 & 0.012 \\
flake8 messages per line & 0.237 & 0.089 & 0.50 & \textless.001 \\
Mean cyclomatic complexity & 3.462 & 3.284 & 0.18 & 0.006 \\
Number of comments & 7.81 & 7.086 & 0.08 & 0.196 \\
Logical lines of code & 46.022 & 43.372 & 0.27 & \textless.001 \\
\end{longtable}

\hypertarget{automatically-generated-code-and-tests}{%
\subsection{Automatically generated code and
tests}\label{automatically-generated-code-and-tests}}

Given the importance of validating AI-generated code using software
tests, we next assessed the ability of GPT-4 to generate tests for its
own code. We first used GPT-4 to generate 20 prompts for each of 5
different areas of research, using the following prompt: ``Please
generate 20 prompts to ask a chatbot to create Python code to solve a
variety of \{statistical and data science, physics, theoretical computer
science, ecology, economics\} problems.''

For each of the generated problems, we created a prompt to generate the
code along with tests for the resulting code, such as the following:

\begin{quote}
Write a Python program to simulate predator-prey interactions using the
Lotka-Volterra equations, given the initial populations, growth rates,
and interaction coefficients. Please embed the code within an explicit
code block, surrounded by triple-backtick markers. Generate realistic
values for any examples, and do not use input() commands. Create code
that is modular and well-commented. Then, generate a set of pytest tests
that exercise each of the functions, embedded in a separate code block.
\end{quote}

\begin{figure}

{\centering \includegraphics{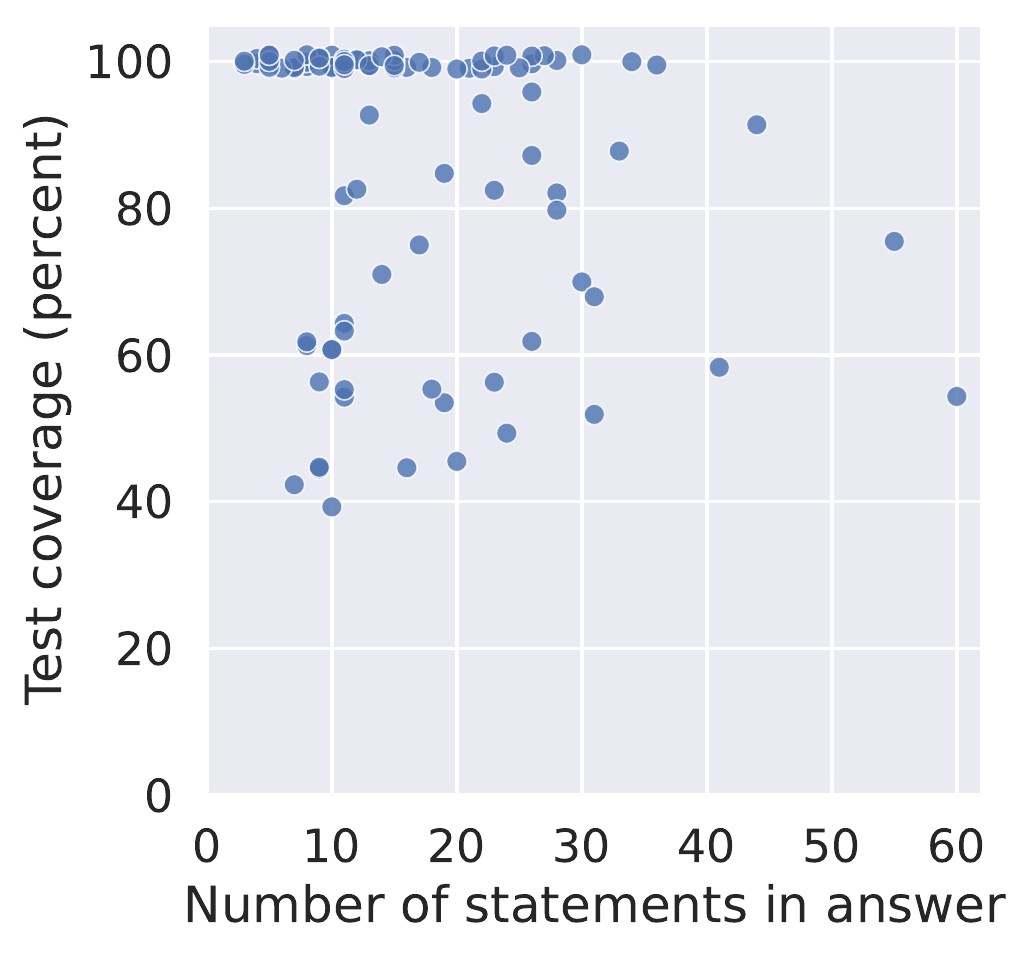}

}

\caption{\label{fig-test-coverage}Percentage of test coverage as a
function of the number of statements in the automatically generated
code.}

\end{figure}

We first examined whether each generated script would execute without
failure; of the 100 generated scripts, 97 executed successfully. We then
examined test coverage using the \texttt{Coverage.py} tool. As shown in
Figure~\ref{fig-test-coverage}, the majority of files had test coverage
of 100\%, with 94\% showing a coverage of at least 50\% ad a minimum
coverage of 40\%. There was a weak but statistically significant
negative relationship between the number of statements and the level of
coverage (Spearman r = -0.23, p = .02). A median of three tests were
generated for each file.

Running the tests required minor automated modification to the code,
since the tests required importing the relevant functions but the names
of the output files were not known to the LLM. After fixing this issue,
45 of the 100 tests completed successfully. The most common source of
test failure was failure of a value assertion (47/100); in these cases,
it was not immediately possible to determine whether the test or code
was incorrect, without additional debugging. In ten cases, the test
failed because the code did not raise the error expected by the test. In
six cases, other errors were raised (Type, Index, Value, and
ZeroDivision errors).

In summary, our analyses of automatic test generation demonstrate that
GPT-4 can successfully generate testing code with good coverage, but
that these tests fail often, requiring additional debugging to determine
the root cause of the test failure.

\hypertarget{conclusions}{%
\subsection{Conclusions}\label{conclusions}}

Our analyses demonstrate that GPT-4 has strong Python code generation
abilities, confirming the results of Bubeck et al. (2023). At the same
time, the prevalence of errors in the generated code suggests that
humans must remain in the loop in order to ensure the accuracy of any
resulting code. Our interactive prompting experiment showed that a
relatively novice prompt engineer can successfully solve coding problems
within a small number of prompts the majority of the time; however, a
sizeable minority of problems would have required significant human
debugging in order to solve. An open question is whether re-prompting in
a new context may have led to more successful outcomes in these cases.

Comparisons of Python code refactored using GPT-4 to the original code
demonstrated that GPT-4 improved the quality of the code, at least as
measured by common metrics of software quality and standards compliance.
It should be emphasized that these results do not assess the accuracy of
the code; rather, they suggest that GPT-4 can help programmers achieve
code that is cleaner and potentially more maintainable than the
original. Given that GPT-4 refactoring did not eliminate all standards
compliance issues, the combination of GPT-4 with other code formatting
tools (such as \texttt{black}) would likely result in even further
improvements.

The examination of test generation by GPT-4 demonstrated that it was
able to generate tests with a high degree of test coverage, but those
tests failed a majority of the time. Such test failures require
additional human effort to diagnose, since it is not immediately clear
whether the failure is due to inaccurate code, and inaccurate test, or
both. These results suggest that while GPT-4 is a very useful tool for
generating testing frameworks for new code, the specific test examples
should be designed and implemented by a human with domain expertise to
ensure that the tests are accurately assessing the intended behavior for
the code.

There has been substantial speculation regarding the continued role of
human programmers in the face of AI coding tools. The present results
suggests that even with the latest generation of AI systems
(i.e.~GPT-4), human involvement is essential to ensure validity and
accuracy of the resulting code. This seems to be especially the case
when programming of mathematical concepts is involved. The lack of
confidence calibration of tools like GPT-4 means that they will present
answers in the same way regardless of the degree of support for the
answer.

The prompts used in the present research are almost certainly
suboptimal, and thus may be underestimating the potential performance of
the model. For example, recent work has shown that chain-of-thought
prompting can substantially improve the perfomance of LLMs on complex
problems requiring reasoning (Prystawski, Thibodeau, and Goodman 2022;
Wei et al. 2023), and this seems to extend to coding as well\footnote{https://martinfowler.com/articles/2023-chatgpt-xu-hao.html}.
Further work is needed to examine the degree to which such improved
prompting techniques might improve the performance of LLMs on complex
coding problems, and at this point our results should be taken as a
lower bound on the performance of these models.

\hypertarget{acknowledgments}{%
\subsection{Acknowledgments}\label{acknowledgments}}

Thanks to Mark Chen, David Coats, and Noah Goodman for helpful comments
and discussion during the development of this work.

\hypertarget{references}{%
\section*{References}\label{references}}
\addcontentsline{toc}{section}{References}

\hypertarget{refs}{}
\begin{CSLReferences}{1}{0}
\leavevmode\vadjust pre{\hypertarget{ref-Bubeck2023}{}}%
Bubeck, Sébastien, Varun Chandrasekaran, Ronen Eldan, Johannes Gehrke,
Eric Horvitz, Ece Kamar, Peter Lee, et al. 2023. {``Sparks of
{Artificial} {General} {Intelligence}: {Early} Experiments with
{GPT}-4.''} arXiv. \url{https://doi.org/10.48550/arXiv.2303.12712}.

\leavevmode\vadjust pre{\hypertarget{ref-Chen2021}{}}%
Chen, Mark, Jerry Tworek, Heewoo Jun, Qiming Yuan, Henrique Ponde de
Oliveira Pinto, Jared Kaplan, Harri Edwards, et al. 2021. {``Evaluating
{Large} {Language} {Models} {Trained} on {Code}.''} arXiv.
\url{https://doi.org/10.48550/arXiv.2107.03374}.

\leavevmode\vadjust pre{\hypertarget{ref-FinnieAnsley2022}{}}%
Finnie-Ansley, James, Paul Denny, Brett A. Becker, Andrew Luxton-Reilly,
and James Prather. 2022. {``The Robots Are Coming: {Exploring} the
Implications of Openai Codex on Introductory Programming.''} In
\emph{Australasian {Computing} {Education} {Conference}}, 10--19.

\leavevmode\vadjust pre{\hypertarget{ref-Fowler2019}{}}%
Fowler, Martin. 2019. \emph{Refactoring: {Improving} the {Design} of
{Existing} {Code}}. Addison-Wesley.

\leavevmode\vadjust pre{\hypertarget{ref-Halstead1977}{}}%
Halstead, Maurice Howard. 1977. \emph{Elements of {Software} {Science}}.
Elsevier.

\leavevmode\vadjust pre{\hypertarget{ref-Ji2023}{}}%
Ji, Ziwei, Nayeon Lee, Rita Frieske, Tiezheng Yu, Dan Su, Yan Xu, Etsuko
Ishii, et al. 2023. {``Survey of {Hallucination} in {Natural} {Language}
{Generation}.''} \emph{ACM Computing Surveys} 55 (12): 1--38.
\url{https://doi.org/10.1145/3571730}.

\leavevmode\vadjust pre{\hypertarget{ref-Prystawski2022}{}}%
Prystawski, Ben, Paul Thibodeau, and Noah Goodman. 2022.
{``Psychologically-Informed Chain-of-Thought Prompts for Metaphor
Understanding in Large Language Models.''} arXiv.
\url{https://doi.org/10.48550/arXiv.2209.08141}.

\leavevmode\vadjust pre{\hypertarget{ref-Wagenmakers2007}{}}%
Wagenmakers, Eric-Jan, Han L. J. Van Der Maas, and Raoul P. P. P.
Grasman. 2007. {``An {EZ}-Diffusion Model for Response Time and
Accuracy.''} \emph{Psychonomic Bulletin \& Review} 14 (1): 3--22.
\url{https://doi.org/10.3758/BF03194023}.

\leavevmode\vadjust pre{\hypertarget{ref-Wei2023}{}}%
Wei, Jason, Xuezhi Wang, Dale Schuurmans, Maarten Bosma, Brian Ichter,
Fei Xia, Ed Chi, Quoc Le, and Denny Zhou. 2023. {``Chain-of-{Thought}
{Prompting} {Elicits} {Reasoning} in {Large} {Language} {Models}.''}
arXiv. \url{https://doi.org/10.48550/arXiv.2201.11903}.

\end{CSLReferences}

\hypertarget{appendix-1-prompts-used-for-interactive-coding-with-gpt-4}{%
\section{Appendix 1: Prompts used for interactive coding with
GPT-4}\label{appendix-1-prompts-used-for-interactive-coding-with-gpt-4}}

Note: The number in brackets denotes the number of prompts necessary to
be judged successful; NS denotes that the problem was abandoned before
success.

\begin{enumerate}
\def\labelenumi{\arabic{enumi}.}
\item
  {[}4{]} Create python code that generates a new class called
  LinearRegressionStats which extends
  sklearn.linear\_model.LinearRegression() to compute the t statistic
  and p-value for each regressor in the model.
\item
  {[}1{]} Create python code to create an abstract base class called
  AbstractPublication to represent publications, with a method called
  from\_pubmed(). Then create a class called Article that inherits
  AbstractPublication. The from\_pubmed() method should take in a pubmed
  record downloaded using the biopython.Entrez tools, and should convert
  the author list to a variable called authors and the title to a
  variable called title. In the main function, perform a pubmed search
  for the query ``cognitive control'' and convert each record into an
  instance of the Publication class, saving them to a dictionary that
  uses the pubmed ID as the key.
\item
  {[}1{]} Create python code to download 1000 abstracts from pubmed
  matching the query ``cognitive control'', clean up the text by
  removing standard English stopwords and lexicalizing the words, and
  apply topic modeling using latent Dirichlet allocation to find 10
  topics. Print the highest scoring words for each topic.
\item
  {[}NS{]} Generate python code to simulate a drift diffusion model of
  response times. Simulate 1000 trials from this model, and estimate the
  drift rate, boundary separation, and starting point parameters using
  the EZ-diffusion model.
\item
  {[}1{]} Create a python application takes in a set of csv files, each
  of which includes two columns: one called ``rt'' that contains a
  response time in miiliseconds, and another ``correct'' that is binary
  (1/0) denoting whether the response was correct or not. there should
  also be another variable, called ``compatible'', that contains binary
  values (0/1) with half of the rows set to zero and half to 1. combine
  these files into a single data frame, using the individual file names
  as an index variable. then perform a linear mixed model that computes
  the effect of the compatible factor, with a random intercept and slope
  for the different file names. also create a test that automatically
  generates a set of data files with random values, and then tests the
  function to assess whether it properly returns the true values.
\item
  {[}7{]} Create python code to find a set of articles matching the
  query ``cognitive control'' using pubmed. For each article extract the
  first and last author and their affilations. Then find their
  institutional web site and scrape their email address from the web
  site.
\item
  {[}1{]} Create python code that uses the github api to download the
  100 most recently committed python files.
\item
  {[}8{]} create python code to load each python file from a directory
  called ``python\_files'', and for each file compute cyclomatic
  complexity, maintainability index, and halsted complexity metrics for
  each file using the radon package.
\item
  {[}2{]} Generate a python script that performs crossvalidation with
  feature selection incorrectly performed outside of the crossvalidation
  loop.
\item
  {[}1{]} Please create code to generate some test data for this code
  (\#9)
\item
  {[}NS{]} Write a python function that takes as input a textfile, then
  create an animation that: {[}1{]} Displays X lines of the textfile at
  a time, with Y characters per line. If the text file has more than Y
  characters in a line, then move to the n ext line.\\
  {[}2{]} Scross down Z lines of the textfile every second, in a
  continuous way. {[}3{]} Animate it in a new popup window. {[}4{]}
  Record the animation for R seconds, and save it as a .avi file. (based
  on a talk by Sebastien Bubeck)
\item
  {[}NS{]} Create python code to implement a three-layer transformer
  model, and train that model using a small text corpus.
\item
  {[}NS{]} Implement python code to simulate a logistic map, and create
  a plot of x versus r.
\item
  {[}NS{]} Create python code to implement the greedy equivalence search
  (GES) algorithm, and code to generate test data that can be used to
  test the implementation.
\item
  {[}1{]} Generate a python class to represent a directed graph.
  generate a function that takes in a directed graph and a two sets of
  vertices and and returns whether those two sets are d-separated. Add a
  test that compares the results from this code to the d-separation
  results computed using the NetworkX library
\item
  {[}2{]} Create code in python to perform a simulation that repeatedly
  generates samples of random variates from 6 different distributions
  (normal, uniform, chi-squared, poisson, exponential, and beta) and
  computes the mean of each sample, saving it for later use. Then,
  generate a figure that shows the distribution of the means for each of
  the distributions.
\item
  {[}NS{]} Create python code to simulate a drift diffusion model with
  collapsing bounds.
\item
  {[}4{]} Generate a python class that fits hurdle regression model,
  with a scikit-learn type interface.
\item
  {[}NS{]} Create a simulation in python that generates a version of the
  ``corridor of stability'' plot by Felix Schonbrodt. The plot should
  show how correlation estimates become less variable as the sample size
  increases.
\item
  {[}1{]} Create a python class to perform spatial smoothing on a nifti
  image using a median filter.
\item
  {[}2{]} Please create a python class that implements the PC causal
  inference algorithm.
\item
  {[}2{]} Generate a python function that takes in a piece of English
  text and performs linguistic analysis on it, returning values for
  sentiment analysis (positive/negative) and for linguistic complexity.
\item
  {[}3{]} Create python code to analyze human performance on the Daw
  two-step reinforcement learning task. the code should take in a data
  frame with human responses on each trial for each of the two steps
  along with the outcomes from each step. it should return the indices
  of model-based and model-free behavior.
\item
  {[}3{]} Create a new crossvalidation class that implements balanced
  cross-validation for regression that follows the scikit-learn class
  structure for crossvalidation objects. this requires finding a set of
  splits that vary minimally in their distributions, which can be
  assessed using an F statistic.
\item
  {[}1{]} Create python code to simulate the effect of overfitting in
  regression. 1) generate synthetic data with 32 observations for two
  variables from a bivariate normal distribution with a correlation of
  0.5. 2) fit three models to the data: a linear regression model, a
  2nd-order polynomial regression model, and a 9-th order polynomial
  model. 3) Compute the error for each of these models on the synthetic
  training data, and on a synthetic test dataset generated from the same
  distribution. 4) plot the fitted lines for each of the fitted models
  overlaid on the training data.
\item
  {[}NS{]} Please create python code to 1) load data from
  https://raw.githubusercontent.com/IanEisenberg/Self\_Regulation\_Ontology/master/Data/Complete\_02-16-2019/meaningful\_variables.csv
  into a data frame, select all variables that begin with
  `upps\_impulsivity\_survey', `sensation\_seeking\_survey',
  `bis11\_survey', or `dickman\_survey', 3) perform exploratory factor
  analysis over these variables using a range of 1 to 5 factors, 4)
  identify which model is preferred based on minimum BIC, and 5) for
  each factor in the preferred solution, display which variables were
  more strongly loaded on that factor versus other factors. Please write
  this in a modular way and output one function at a time.
\item
  {[}NS{]} Generate python code to obtain daily weather reports from
  palo alto, california for 1960 to 2000. from these data, compute the
  maximum temperature in each month, and plot the timeseries of maximum
  monthly temperatures over that time period.
\item
  {[}1{]} Generate python code to read in an image and render the image
  using ascii art.
\item
  {[}3{]} Create python code to take a document written in Markdown and
  render it as a PDF, with an image for the header. please use argparse
  for the input values
\item
  {[}1{]} Generate a class in the scikit-learn style to perform
  principal component analysis using the power iteration method. the
  algorithm should be coded from scratch rather than using an external
  library.
\item
  {[}1{]} Create a python wrapper for the statsmodels ttest\_ind
  function, which takes the same arguments and returns a report similar
  to the one returned by the t.test function in R
\item
  {[}1{]} Create a simulation to demonstrate the use of randomization to
  obtain a null distribution for hypothesis testing. First, generate a
  bivariate dataset with a sample size of 100 and a correlation of 0.1
  between the two variables. Compute the correlation coefficient and
  obtain a p-value against the null hypothesis of r=0. Then, randomly
  shuffle one of the variables 5000 times and compute the correlation
  each time. Compute an empirical p-value for the actual R value based
  on the null distribution.
\end{enumerate}

\end{document}